\def\year{2018}\relax
\documentclass[letterpaper]{article}
\usepackage{aaai18}  
\usepackage{times}  
\usepackage{helvet}  
\usepackage{courier}  
\usepackage{url}  
\usepackage{graphicx}  
\usepackage{nameref}
\usepackage{graphicx}
\usepackage[utf8]{inputenc}
\usepackage{graphicx}
\usepackage{amsfonts,amssymb}
\usepackage{multirow}
\usepackage{mathtools}
\usepackage{algpseudocode} 

\usepackage{algorithm} 
\usepackage[dvipsnames]{xcolor}
\newtheorem{example}{Example}
\newcommand{\pop}[1]{\Pi_{#1}}
\newcommand{\lv}[1]{#1}
\newcommand{\lvs}[1]{\mathbf{#1}}
\newcommand{\ind}[1]{#1}
\newcommand{\inds}[1]{\mathbf{#1}}

\newcommand{\pred}[1]{\ensuremath{\mathsf{#1}}}
\newcommand{\true}{\ensuremath{\mathsf{True}}}

\newcommand{\atoms}[1]{\ensuremath{\mathcal{#1}}}
\newcommand{\grounds}[1]{\ensuremath{\mathcal{G(#1)}}}
\newcommand{\WF}[2]{\ensuremath{\left\langle #1, #2 \right\rangle}}
\newcommand{\counterTrue}[2]{\ensuremath{\eta_T(#1,#2)}}
\newcommand{\counterFalse}[2]{\ensuremath{\eta_F(#1,#2)}}

\newcommand{\pa}{\mathbf{parents}}
\newcommand{\bfc}{\mathbf{c}}
\newcommand{\bfx}{\mathbf{x}}
\newcommand{\bfw}{\mathbf{w}}
\usepackage{tabularx}

\newcommand{\minimize}{\operatornamewithlimits{\mathrm{minimize}}}

\newcommand{\brlr}{\textsc{b-rlr}}
\newcommand{\ilprlr}{\textsc{ilp-rlr}}
\newcommand{\mlnb}{\textsc{mln-b}}
\newcommand{\agglr}{\textsc{agg-lr}}
\algnewcommand{\LeftComment}[1]{\State \(\triangleright\) #1}
\algnewcommand{\func}[1]{\textsc{#1}}

\begin{document}
\title{Structure Learning for Relational Logistic Regression:\\
An Ensemble Approach}
\author{
Nandini Ramanan \\
Indiana University, Bloomington
\And
Gautam Kunapuli \\
The University of Texas at Dallas 
\And
Tushar Khot \\
Allen Institute for Artificial Intelligence \\
\AND
Bahare Fatemi \\
University of British Columbia
\And  
Seyed Mehran Kazemi \\
University of British Columbia
\And
David Poole \\
University of British Columbia
\And
Kristian Kersting \\
TU Darmstadt University
\AND
Sriraam Natarajan \\
The University of Texas at Dallas}

\maketitle
\begin{abstract}
We consider the problem of learning Relational Logistic Regression (RLR). Unlike standard logistic regression, the features of RLRs are first-order formulae with associated weight vectors instead of scalar weights. We turn the problem of learning RLR to learning these vector-weighted formulae and develop a learning algorithm based on the recently successful functional-gradient boosting methods for probabilistic logic models. We derive the functional gradients and show how weights can be learned simultaneously in an efficient manner. Our empirical evaluation on standard and novel data sets demonstrates the superiority of our approach over other methods for learning RLR.
\end{abstract}

\section{Introduction}
Statistical Relational Learning models (SRL)~\cite{GetoorTaskar07,DeRaedtEtAl16} combine the representational power of logic with the ability of probability theory specifically, and statistical models in general to model noise and uncertainty. They have generally ranged from directed models~\cite{KerstingDeRaedt07-BLP,koller1999probabilistic,heckerman2007probabilistic,rlr,NevilleJensen07-RDN,relnn} to undirected models~\cite{richardson2006markov,taskar2007relational,kimmig2012short}. We consider the more recent, well-understood directed model of \textbf{Relational Logistic Regression} (RLR)~\cite{rlr,KazemiEtAl14-RLR,FatemiKP16-LearningRLR}. One of the key advantages of RLR is that they scale well with population size unlike other methods such as Markov Logic Networks~\cite{Poole2014} and hence can potentially be used as a powerful modeling tool for many tasks.

While the models are attractive from the modeling perspective, learning these models is computationally intensive. This is due to the fact that (like the field of Inductive Logic Programming) learning occurs at multiple levels of abstraction, that of the level of objects, sub-group of objects and relations and possibly at the individual instances of the objects. Hence, most methods for learning these models have so far focused on the task of learning the so-called parameters (weights of the logistic function) where the rules (or relational features) are provided by the human expert and the data is merely used to learn the parameters.

We consider the problem of full model learning, also known as \textbf{structure learning} of RLR models. A simple solution to learning these models could be to learn the rules separately using a logic learner and then employ the parameter learning strategies~\cite{huynh2008discriminative}. While reasonably easy to implement, the key issue is that the disconnect between rule and parameter learning can result in poor predictive performance as shown repeatedly in the literature~\cite{natarajan2012gradient,richardson2006markov}. Inspired by the success of non-parametric learning methods for SRL models, we develop a learning method for full model learning of RLR models. 

More specifically, we develop a gradient-boosting technique for learning RLR models. We derive the gradients for the different weights of RLR and show how the rules of the logistic function are learned simultaneously with their corresponding weights. Unlike the standard adaptations of the functional gradients, RLR requires learning a different set of weights per rule in each gradient step and hence requires learning multiple weights jointly for a single rule. As we explain later, the gradients correspond to a set of vector weighted clauses that are learned in a sequential manner. We derive the gradients for these clauses and illustrate how to optimize them. 

Each clause can be seen as a \textbf{relational feature} for the logistic function. We also note that RLR can be viewed as a probabilistic combination function in that it can stochastically combine the distributions due to different set of parents (in graphical model terminology). Hence, if our learning technique is employed in the context of learning joint models, our work can be seen as a new interpretation of learning boosted Relational Dependency Networks (RDNs)~\cite{NevilleJensen07-RDN,natarajan2012gradient}, where the standard aggregators are replaced with a logistic regression combination function which could potentially yield interesting insights into directed SRL models. We demonstrate the effectiveness of this combination function on real data sets and compare against several baselines including the state-of-the-art MLN learning algorithms.

The rest of the paper is organized as follows: first we introduce the necessary background and introduce the notations. Next we derive the gradients and present the algorithm for learning RLR models. Finally, we conclude the paper by presenting our extensive experimental evaluations on standard SRL data sets and a NLP data set and outlining the areas for future research.


\section{Background and Notation}
In this section, we define our notation and provide necessary background for our readers to follow the rest of the paper. Throughout the paper, we assume $True$ is represented by $1$ and $False$ is represented by $0$.

\subsection{Logistic Regression}
Let $Q$ be a Boolean random variable with range \{1, 0\} whose value depends on a set $\{X_1, X_2, \dots, X_n\}$ of random variables. Logistic regression \cite{mccullagh1984generalized} defines the conditional probability of $Q$ given $X_1, X_2, \dots, X_n$ as the sigmoid of a weighted sum of $X_i$s:
\begin{equation}
    Prob(Q=1\mid X_1, \dots, X_n) = \sigma(w_0 + w_1 X_1 + \dots + w_n X_n)
\end{equation}
where $\sigma(z)=1 / (1+\exp(-z))$ is the sigmoid function.

\subsection{Finite-Domain, Function-Free, First-Order Logic}
A \textbf{population} is a finite set of objects. We assume for every object, there is a unique \textbf{constant} denoting that object. A \textbf{logical variable (logvar)} is typed with a population. A \textbf{term} is a logvar or a constant. We show logvars with lower-case letters (e.g., $\lv{x}$) , constants with upper-case letters (e.g., $X$), the population associated with a logvar $\lv{x}$ with $\pop{x}$, and the size/cardinality of the population with $|\pop{x}|$. A lower-case letter in bold represents a tuple of logvars (e.g., $\bfx$), and an upper-case letter in bold is a tuple of constants (e.g., $\mathbf{X}$).

An \textbf{atom} is of the form $\pred{Q}(t_1, \dots, t_k)$ where  $\pred{Q}$ is a functor and each $t_i$ is a term. When $range(\pred{Q})=\{1, 0\}$, $\pred{Q}$ is a predicate. A \textbf{substitution} is of the form $\theta=\{\langle\lv{x_1}, \dots, \lv{x_k}\rangle/ \langle t_1, \dots, t_k\rangle\}$ where $\lv{x_i}$s are logvars and $t_i$s are terms. A \textbf{grounding} of an atom with logvars $\lv{x_1}, \dots, \lv{x_k}$ is a substitution $\{\langle\lv{x_1}, \dots, \lv{x_k}\rangle \slash \langle \ind{X_1}, \dots, \ind{X_k}\rangle\}$ mapping each of its logvars to a constant in the population of the logvar. For a set $\atoms{A}$ of atoms, $\grounds{A}$ represents the set of all possible groundings for the atoms in $\atoms{A}$. A \textbf{literal} is an atom or its negation. A \textbf{formula} $\varphi$ is a literal, the conjunction of two formulae $\varphi_1 \wedge \varphi_2$, or a disjunction of two formulae $\varphi_1 \vee \varphi_2$. The application of a substitution $\theta=\{\langle \lv{x_1}, \dots, \lv{x_k} \rangle \slash \langle t_1, \dots, t_k \rangle\}$ on a formula $\varphi$ is represented as $\varphi\theta$ and replaces each $\lv{x_i}$ in $\varphi$ with $t_i$. An \textbf{instance} of a formula $\varphi$ is obtained by replacing each logvar $\lv{x}$ in $\varphi$ by one of the objects in $\pop{x}$. A \textbf{conjunctive formula} contains no disjunction. 
A \textbf{weighted formula (WF)} is a triple $\WF{\varphi}{w_T, w_F}$ where $\varphi$ is a formula and $w_T$ and $w_F$ are real numbers.

\subsection{Relational Logistic Regression}
Let $\pred{Q}(\lvs{x})$ be a Boolean atom whose probability depends on a set $\atoms{A}$ of atoms such that $\pred{Q} \notin \atoms{A}$. We refer to $\atoms{A}$ as the parents of \pred{Q}. Let $\psi$ be a set of WFs containing only atoms from $\atoms{A}$, $J$ be a function from groundings in $\grounds{A}$ to truth values, and $\theta=\{\lvs{x} \slash \inds{X}\}$ be a substitution from logvars in $\lvs{x}$ to constants in $\inds{X}$. {\bf Relational logistic regression (RLR)} \cite{rlr} defines the probability of $\pred{Q}(\inds{X})$ given $J$ as follows:

\begin{align}\label{RLR-EQ}
&Prob_\psi(\pred{Q}(\inds{X})=1\mid J)\nonumber\\& =\sigma\left(w_0+\sum_{\WF{\varphi}{w_T, w_F}\in \psi}{w_T\cdot\counterTrue{\varphi\theta}{J}+w_F\cdot\counterFalse{\varphi\theta}{J}}\right)
\end{align}
where $w_0$ is a bias/intercept, $\counterTrue{\varphi\theta}{J}$ is the number of instances of $\varphi\theta$ that are true with respect to $J$, and $\counterFalse{\varphi\theta}{J}$ is the number of instances of $\varphi\theta$ that are false with respect to $J$. Note that $\counterTrue{\true}{J}=1$. Also note that the bias can be considered as a WF whose formula is $True$. Following Kazemi et al. \cite{rlr}, without loss of generality we assume the formulae of all WFs for RLR models are conjunctive.
\begin{example}
Let $\pred{Active}(\lv{p})$, $\pred{AdvisedBy}(\lv{s},\lv{p})$, and $\pred{PhD}(\lv{s})$ be three atoms  representing respectively whether a professor is active, whether a student is advised by a professor, and whether a student is a PhD student. Suppose we want to define the conditional probability of $\pred{Active}(\lv{p})$ given the atoms $\atoms{A}=\{\pred{AdvisedBy}(\lv{s},\lv{p}), \pred{PhD}(\lv{s})\}$. Consider an RLR model with an intercept of $-3.5$ which uses only the following WF to define this conditional probability:
\begin{align*}
\psi=\{ \WF{\pred{AdvisedBy}(\lv{s},\lv{p})\wedge \pred{PhD}(\lv{s})}{1,0}\}     
\end{align*}
According to this model, for an assignment $J$ of truth values to \grounds{A}:
\begin{align*}
&Prob_\psi(\pred{Active}(\ind{P})=1 \mid J) =\\& \sigma(-3.5 + 1 \cdot \counterTrue{\pred{AdvisedBy}(\lv{s},\ind{P})\wedge \pred{PhD}(\lv{s})}{J}),    
\end{align*}
where $\counterTrue{\pred{AdvisedBy}(\lv{s},\ind{P})\wedge \pred{PhD}(\lv{s})}{J})= \#\ind{S} \in \pop{s}$ s.t. $\pred{AdvisedBy}(\ind{S},\ind{P}) \wedge \pred{PhD}(\ind{S})$ according to $J$, corresponding to the number of PhD students advised by $\ind{P}$. When this count is greater than or equal to 4, the probability of $\ind{P}$ being an active professor is closer to one than zero; otherwise, the probability is closer to zero than one. Therefore, this RLR model represents ``a professor is active if the professor advises at least 4 PhD students''.
\end{example}
With this background on Relational Logistic Regression, we introduce the Functional Gradient Boosting paradigm in the following section. This enables us to formulate a learning problem for RLR in which we learn both the structure and the parameters simultaneously.

\subsection{Functional Gradient Boosting}\label{fgbSec}
We discuss \textbf{functional gradient boosting} (FGB) approach in the context of relational models. This approach is motivated by the intuition that finding many rough rules-of-thumb of how to change one's probabilistic predictions locally can be much easier than finding a single, highly accurate model. Specifically, this approach turns the problem of learning relational models into a series of \textbf{relational function approximation} problems using the ensemble method of \textbf{gradient-based boosting}. This is achieved by the application of Friedman's~\cite{friedman01} gradient boosting to SRL models. That is, we represent the conditional probability distribution as a weighted sum of regression models that are grown via a stage-wise optimization~\cite{natarajan2012gradient,khot2011learning}. 

The conditional probability of an example $y_i$\footnote{We use the term example to mean the grounded target literal. Hence $y_i = 1$ denotes that the grounding $\pred{Q}(\inds{X})=1$ i.e., the grounded target predicate is true. Following standard Bayesian networks terminology, we denote the parents $\atoms{A}(\pred{Q})$ to include the set of formulae $\psi$ that influence the current predicate $\pred{Q}$.} depends on its parents $\bfx_i \, = \, \pa(y_i)$. The goal of learning is to fit a model $Prob(y \, | \, \mathbf{x}) \propto e^{\psi(y,\mathbf{x})}$, and can be expressed non-parametrically in terms of a potential function $\psi(y_i; \, \bfx_i)$:
\begin{equation}
    Prob(y_i \, | \bfx_i) \, = \, \displaystyle{\frac{e^{\psi(y_i; \, \bfx_i)}}{\sum_{y^\prime} \, e^{\psi(y^\prime; \, \bfx_i)}}}
\end{equation}


At a high-level, we are interested in successively approximating the function $\psi$ as a sum of weak learners, which are relational regression clauses, in our setting. Functional gradient ascent starts with an initial potential $\psi_0$ and iteratively adds gradients $\Delta_i$. After $m$ iterations, the potential is given by $\psi_m = \psi_0 + \Delta_1 + ... + \Delta_m$. Here, $\Delta_m$ is the \textbf{functional gradient} at episode $m$ and is
\begin{equation}\
\Delta_m = \eta_m \cdot E_{\bfx,y}\left[\frac{\partial}{\partial{\psi_{m-1}}} \, \log \,P(y \mid \bfx; \, \psi_{m-1}) \right],
\label{eq: funcgradient}
\end{equation}
where $\eta_m$ is the learning rate. Dietterich {\it et al.}~\cite{dietterich04} suggested evaluating the gradient at every position in every training example and fitting a regression tree to these derived examples i.e., fit a regression tree $h_m$ on the training examples $[(x_i,y_i), \Delta_m(y_i;x_i)]$. They point out that although the fitted function $h_m$ is not exactly the same as the desired $\Delta_m$, it will point in the same direction (assuming that there are enough training examples). Thus, ascent in the direction of $h_m$ will approximate the true functional gradient.

Note that in the functional gradient presented in (\ref{eq: funcgradient}), the expectation $E_{\bfx,y}$ cannot be computed as the joint distribution $P(\mathbf{\bfx}, \, y)$ is unknown.
Instead of computing the functional gradients over the potential function, they are instead computed \textbf{pointwise} for each labeled training example $i$: $\langle \bfx_i, \, y_i \rangle$. Now, this set of local gradients become the training examples to learn a weak regression model that approximates the gradient $\Delta_m$ at stage $m$. 

\begin{figure*}[t]
\centering
\includegraphics[scale=0.36]{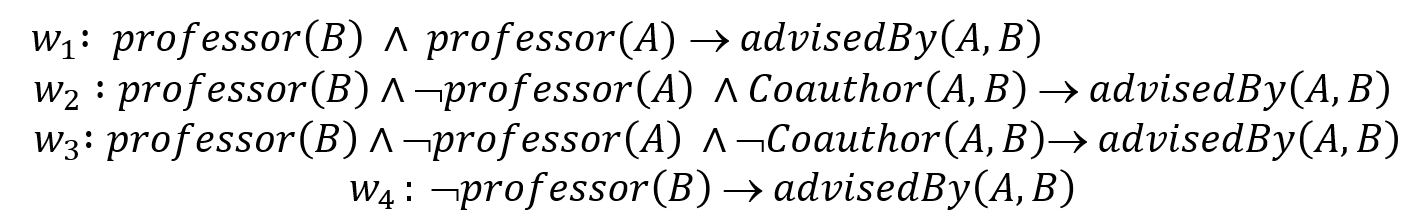}
\caption{Example of an RRC. The task was to predict if $A$ is \pred{AdvisedBy} $B$, given the relations of people at a university.}
\label{fig: RRT}
\end{figure*}

\begin{figure*}[t]
\centering
\includegraphics[scale=0.60]{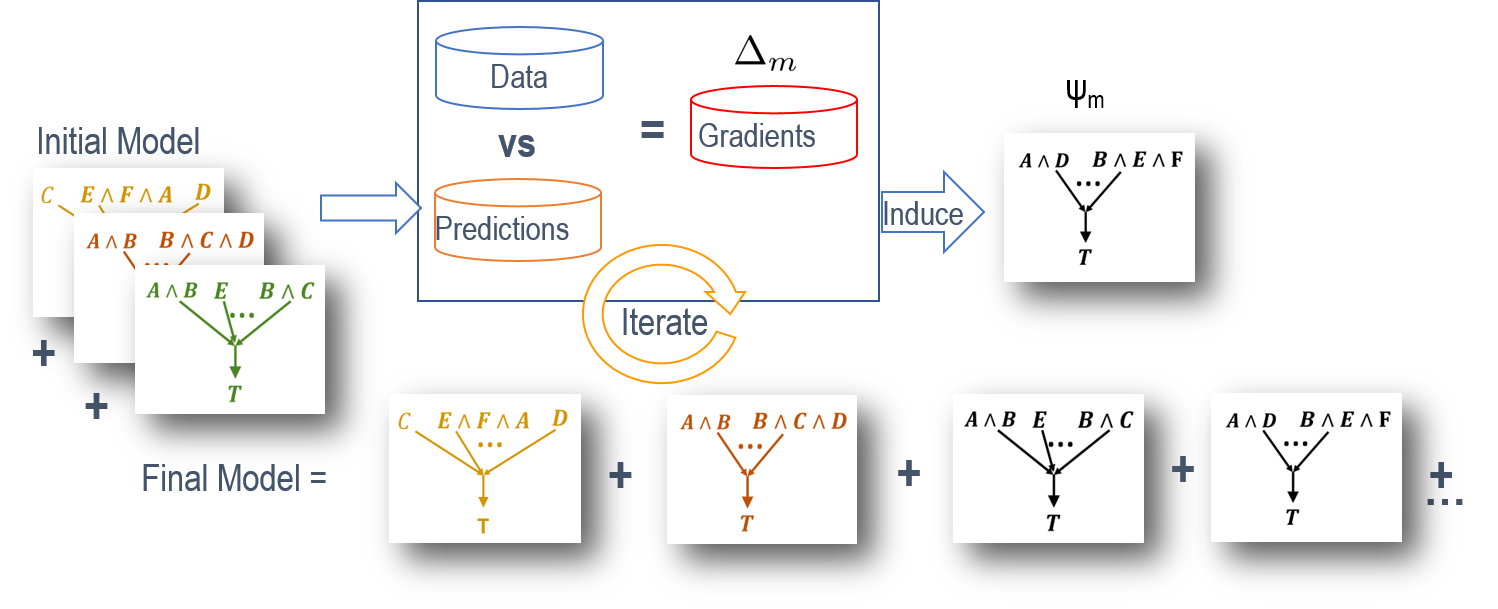}
\caption{Relational Functional Gradient Boosting. This is similar to standard functional gradient boosting (FGB) where trees are induced stage-wise; the key difference is that these trees are relational regression clauses (RRCs).}
\label{fig: RFGB}
\end{figure*}

The functional gradient with respect to $ \psi(y_i=1; \, \bfx_i)$ of the likelihood for each example $\langle y_i, \mathbf{x}_i \rangle$ can be shown to be:
\begin{equation}
\frac{\partial \log Prob(y_i;\, \bfx_i)}{\partial \psi(y_i=1; \, \bfx_i)} =  \mathbb{I}(y_i=1;\bfx_i)- P(y_i=1;\bfx_i)
\end{equation}
where $\mathbb{I}$ is the indicator function, that is $1$, if $y_i = 1$, and $0$ otherwise. This expression is simply {\em the adjustment required to match the predicted probability with the true label of the example}. If the example is positive and the predicted probability is less than $1$, this gradient is positive indicating that the predicted probability should move towards $1$. Conversely, if the example is negative and the predicted probability is greater than 0, the gradient is negative driving the value the other way. 

This elegant gradient expression might appear simple, but in fact, naturally and intuitively captures, example-wise, the general direction that the overall model should be grown in. The $I-P$ form of the functional gradients is a consequence of the sigmoid function as a modeling choice, and is a defining characteristic of FGB methods. As we show below, our proposed approach to RLR also has a similar form. The significant difference, however, is in the novel definition of the potential function $\psi$.

In prior work,  \textbf{relational regression trees} (RRTs)~\cite{tilde} were used to fit the gradient function $\Delta_m$ to the pointwise gradients for every training example. Each RRT can be viewed as defining several new feature combinations, one corresponding to each path from the root to a leaf. A key difference in our work is that we employ the use of weighted formulae (vector-weighted clauses\footnote{We use formulae and clauses interchangeably from hereon.}, to be precise) as we explain later. From this perspective, our work is closer to the boosting MLN work that employed the use of weighted clauses. We generalize this by learning a weight vector per clause that allows for a more compact representation of the true and false instances of the formula. An example of a weighted clause is provided in Figure~\ref{fig: RRT} where there are four clauses for predicting {\em advisedBy(A,B)}. Note that while we show the standard weighted clauses similar to a MLN, our weighted clauses have an important distinction - corresponding to each clause is a weight vector instead of a single scalar weight which captures the weight of true groundings, false groundings and the uninformed prior weights of the clause. The gradient-boosting that we develop in the next section builds upon these clauses and as mentioned earlier is similar to MLN boosting with the key difference being that instead of learning one weight per clause, we learn three weights in the vector.

The key intuition with boosting regression clauses is that each clause will define a new feature combination and the different clauses together capture the {\em latent relationships} that are learned from the data. While the final model itself is linear (as it is the sum of the weighted groundings of all the clauses), the clauses themselves define richer features thus allowing for learning a more complex model than a simple linear one. Figure~\ref{fig: RFGB} presents the schematic for boosting. The idea is that first a regression function (shown as a set of clauses) is learned from the training examples and these clauses are then used to determine the gradients (weights) of each example in the next iteration. The gradient is typically computed in the prior work as $I-P$. Once the examples are weighted, a new set of clauses are induced from them. These clauses are then considered together and the {\em regression values are added} when weighing the examples and the process is iterated.




There are several benefits of the boosting approach for learning RLR models. First, being a non-parametric approach (i.e., the model size is not chosen in advance), the number of parameters naturally grows as the number of training episodes increases.  In turn, relational features as clauses are introduced only as necessary, so that a potentially large search space is not explicitly considered.  Second, such an algorithm is fast and straightforward to implement.  One could potentially employ any relational regression learner in the inner loop to learn several types of models. Third, as with previous relational models, the use of boosting for learning RLR models makes it possible to \textbf{learn the structure and parameters simultaneously} making them an attractive choice for learning from large scale data sets~\cite{malec2016inductive,yang2017combining}.

\section{Functional Gradient Boosting for RLR}

\subsection{Preliminaries}

Given the background on RLR and the gradient-boosting approach, we now focus on the learning task of RLR. Let us rewrite the conditional probability of an example $y$ given weighted formulae $\WF{\varphi_1}{ w_{T1}, w_{F1}}, \cdots, \WF{\varphi_k}{ w_{Tk}, w_{Fk}}$ corresponding parents 
$J_1, \cdots, J_k$ in the RLR model as:
\begin{align}
    Prob({\bf y}&=1 \,| \, \pred{J}_1,\cdots,\pred{J}_k)= \nonumber\\&\sigma ( w_0 \nonumber
    +w_{T1}  \counterTrue{\varphi_1\theta}{J_1} + w_{F1} \cdot \counterFalse{\varphi_1\theta}{J_1} \nonumber\\
    &+\cdots\nonumber\\
    &+w_{Tk}  \counterTrue{\varphi_k\theta}{J_k} + w_{Fk} \cdot \counterFalse{\varphi_k\theta}{J_k})
\label{rlrexpanded}
\hspace{-0.1in}
\end{align}
where $\sigma(\cdot)$ is the sigmoid function. For example, let $y=\pred{Popularity}(a)$ indicate the popularity of a professor $a$. Consider two formulae $\phi_1$=$\pred{Publication}(A, P)$ and $\phi_2$=$\pred{AdvisedBy(A, S)}$. The weights of the first formula control the influence of the number of publications on the popularity of the professor where $J_1=\pred{Publication}(a, P)$. Similarly the second formula controls the influence of the number of students advised by the professor. For learning a model for RLR, we thus need to learn these clauses $\phi_i$ and their weights $w_{Ti}, w_{Fi}$ (the parents are determined by the structure of the clause). Also, we can assume that the bias term $w_0$ can be part of the weight vectors for all the learned clauses. This allows a greedy approach that incrementally adds new clauses, such as FGB, to automatically update the bias term by learning $w_0$ for each new clause.  Our learning problem can be defined as:\\



\noindent \textsf{{\bf Given:} A set of grounded facts (features) and the corresponding 
positive and negative grounded literals (examples)}\\
\noindent \textsf{{\bf To Do:} Learn the set of formulae $\varphi_i$ with their corresponding weight vector $\bfw_i=[w_0, w_1, w_2]$}.\\ 

To simplify the learning problem, we introduce \textbf{vector-weighted clauses} (formulae), denoted as \{$\bfw: \pred{Clause}$\}, that are a generalization of traditional weighted clauses with single weights. More specifically, our weighted clauses employ \textbf{three dimensions}, that is $\bfw = [w_0, \, w_1, \, w_2]^T$, where $w_0$ is a bias/intercept, $w_1$ is the weight over the satisfiable groundings of the current clause (analogous to $w_{Ti}$) and $w_2$ is the weight of the unsatisfiable groundings of the current clause (analogous to $w_{Fi}$). 
We also use a short hand notation $t_i$ and $f_i$ for the two grounding counts $\counterTrue{\varphi_1\theta}{J_1}$ and $\counterFalse{\varphi_1\theta}{J_1}$ respectively in Equation~\ref{rlrexpanded}.

\begin{example}
Consider an RLR model for defining the conditional probability of $y=\pred{Popularity}(a)$ which has only one WF: 
\begin{align*}
 &\WF{\pred{Publication}(A, B)}{w_T, w_F} \equiv\\ & [w_0, w_1, w_2]: \pred{Popularity}(A) \coloneq \pred{Publication}(A, B)
\end{align*}

Let $t_y \, = \, \sum_b \, \pred{Publication}(a, \, b)$ be the number of instances of $b$ for which $\pred{Publication}(a, \, b)$ is true for the current grounding of $y$, and let $f_y \, = \, \sum_b \, (1 - \pred{Publication}(a, \, b))$ be the number of instances of $b$ for which $\pred{Publication}(a, \, b)$ is false for the example $y$. Using vector-weighted clauses in the RLR model, we can compute 
\begin{align*}
    Prob(\pred{Popularity}(a) \, |& \, \pred{Publication}(a, B)) \, =\\& \, \sigma \left( w_0 \, + \, w_1 \cdot t_y \, + \, w_2 \cdot f_y \right), \,\,  \forall a. 
\end{align*}

\end{example}

\subsection{RFGB for RLR}

Our goal is to learn the full structure of the model, which involves learning two concepts -- the structure (formulae/clauses) and their associate parameters (the weight vectors). To adapt functional gradient boosting to the task of learning RLR, we map this probability definition over the parameter space $w_0, w_1, w_2$ to the functional space, $\psi$:
\begin{equation}
\begin{aligned}
    P(y_i \, = \, 1 \, | \, \bfx_i) \, &= \, \sigma (\psi \left(y_i; \, \bfx_i \right)) \, \\&= \,  \sigma \left(w_0 + w_1 \cdot t_i + w_2 \cdot f_i \right)
\end{aligned}
\end{equation}

Recall that in FGB, the gradients of the likelihood function with respect to the potential function are computed separately for each example. Correspondingly, the regression function $\psi$ for the $i$-th example needs to be clearly defined and is:
\begin{equation}
    \psi(y_i; \, \bfx_i) \, = \, w_0 \, + \, w_1 \cdot t_i \, + \, w_2 \cdot f_i.
\end{equation}
The key difference to the existing gradient boosting methods for RDNs~\cite{natarajan2012gradient} and MLNs~\cite{khot2011learning} is that the RLR learning algorithm needs to learn a weight vector per clause instead of a single weight. 

Also, recall that while in traditional parametric gradient-descent, one would compute the parametric gradient over the loss function and iteratively update the parameters with these gradients, for gradient boosting, we first compute the functional gradients over the log-likelihood given by:
\begin{equation}
    \Delta(y_i; \, \bfx_i) \, = \, \mathbb{I}(y_i \, = \, 1) \, - \, P(y_i \, = \, 1; \, \bfx_i)
\end{equation}
where $\mathbb{I}$ is the indicator function. As with other relational functional gradients (see Section \nameref{fgbSec}), this elegant expression naturally falls out when the log-likelihood of the sigmoid is differentiated with respect to the function. 
As before, the gradient is simply the adjustment required for the probabilities to match the observed value ($y_i$) in the training data for each example. Note that this is simply the outer gradient, that is, the gradient is computed for each example and a single vector-weighted clause needs to be learned for this set of gradients. While learning the clause itself, we must optimize a different loss function as we show next. 

In order to generalize beyond the training examples, we fit a regression function $\psi$ (which is essentially a {\em vector-weighted clause}) over the training examples such that the \textbf{squared error} between $\psi(y_i; \, \bfx_i)$ and the functional gradient $\Delta(y_i; \, \bfx_i)$ is minimized over all the examples. The inner loop thus amounts to learning vector-weighted clauses 
such that we minimize the (regularized) squared error between the RLR model and the functional gradients over the $n$ training examples:
\begin{equation}
\begin{aligned}
   \minimize_{w_0, w_1, w_2} \, \sum_{i=1}^n \, &(w_0 \, + \, w_1 t_i \,+ \, w_2 f_i \,\, - \,\,  \Delta(y_i; \, \bfx_i))^2 \, \\&+ \, \lambda \, (w_0^2 + w_1^2 + w_2^2),
   \label{eq: BRLR minimization}
\end{aligned}
\end{equation}
where $\lambda > 0$ is a regularization parameter. In principle, $\lambda$ can be chosen using a line search with a validation set when the size of the data sets are large. However, in our data sets, we only considered a few $\lambda$ values from the set of $\{10^2, 10^{2.5}, 10^3, 10^{3.5}\}$ and chose to present the best $\lambda$ corresponding to the test set. 

Close inspection of the loss function above reveals that solving this optimization problem amounts to fitting \textbf{count features}: $\bfc_i \, = \, [1, \, t_i, \, f_i]$ for each grounded example $i$ to the corresponding functional gradient, $\Delta_i$. Note that the equation (\ref{eq: BRLR minimization}) can be viewed as a regularized least-squares regression problem to identify weights $\bfw = [w_0, \, w_1, \, w_2]^T$. The problem (\ref{eq: BRLR minimization}) can be written in vector form as 
\begin{equation*}
 \min_{\bfw} \, \| C \bfw\, - \, \Delta\|^2 \, + \, \lambda \|\bfw\|^2   
\end{equation*}
where the $i$-th row of the count matrix $C$ are the count features $\bfc_i$ of the $i$-th example. This problem has a closed-form solution that can be computed efficiently:
\begin{equation}
    \bfw \, = \, (C^T C \, + \, \lambda I)^{-1} \, C^T \Delta.
    \label{eq: BRLR minimization solution}
\end{equation}
The quantity $C^T C$ captures the \textbf{count covariance} across examples, while the quantity $C^T \Delta$ captures the \textbf{count-weighted gradients}:
\begin{equation}
\begin{aligned}
&C^T C \, = \, \left[ \begin{array}{ccc}
n & \sum_{i=1}^n t_i & \sum_{i=1}^n f_i \\ [3pt]
\sum_{i=1}^n t_i & \sum_{i=1}^n t_i^2 & \sum_{i=1}^n t_i f_i \\ [3pt]
\sum_{i=1}^n f_i & \sum_{i=1}^n t_i f_i & \sum_{i=1}^n f_i^2
\end{array} \right], \,\,\, \\&     
C^T \Delta \, = \, \left[ \begin{array}{c}
\sum_{i=1}^n \Delta_i  \\ [3pt] 
\sum_{i=1}^n t_i \Delta_i  \\ [3pt]
\sum_{i=1}^n f_i \Delta_i
\end{array} \right].
\end{aligned}
\end{equation}

In this manner, functional gradient boosting enables a natural combination of conditionals over all the examples. This weight update forms the backbone of our approach: boosted relational logistic regression or \brlr. 

\subsection{Algorithm for \brlr}

We outline the algorithm for boosted RLR (\brlr) learning in Algorithm~\ref{algo:brlr}. We initialize the regression function with an uniform prior $\gamma$ i.e. $F_0(y_i) = \gamma$ (line 2). Given the input training examples $Y$ which correspond to the grounded instances of the target predicate $y$ and the set of facts, i,e., the grounded set of all other predicates (denoted as $X$) in the domain, the goal is to learn the set of vector-weighted clauses that influence the target predicate. 

Since there could potentially be multiple target predicates (when learning a joint model such as RDN where each influence relation is an RLR), we denote the current predicate as $p$. In the $m^{th}$ iteration of  functional-gradient boosting, we compute the functional gradients for these examples using the current model $F_m$ and the parents of $\bf y$ as per this model (line 7). Given these regression examples $S_p$, we learn a vector-weighted clause using \func{FitRegression}. This function uses all the other facts $X$ to learn the structure and parameters of the clause. We then add this regression function, $\hat{\psi}_m$ approximating the functional gradients to the current model, $F_m$. We repeat this over $M$ iterations where $M$ is typically set to 10 in our experiments. 

\begin{algorithm}
\begin{algorithmic}[1]
\Function{\brlr}{$Y$, $X$, $p$}
\State $F_0$ := $\gamma$
\For {$1 \leq m \leq M$} \Comment{M gradient steps}
\State $F_{m} := F_{m-1}$
\LeftComment{Compute gradients, $\Delta_i$ for $y_i \in Y_p$}
\State $S_p$ := \func{ComputeGradients}($Y_p$, $X$, $F_m$)
\State $\hat{\psi}_{m}$ := \func{FitRegression}($S_p$, $X, Y_p$) \Comment{Learn vector-weighted regression clause}
\State $F_{m} := F_{m} + \hat{\psi}_{m}$  \Comment{Update model}
\EndFor
\State \Return $F_m$
\EndFunction
\end{algorithmic}
\caption{Boosted Relational Logistic Regression (\brlr) learning}
\label{algo:brlr}
\end{algorithm}

Next, we describe \func{FitRegression} to learn vector-weighted clauses from input regression examples $S$, facts $D$ and target predicate \pred{p}$(x)$ in Algorithm~\ref{algo:fitReg}. We initialize the vector-weighted clause with empty body and zero weights i.e. $[0, 0, 0]^T: y \coloneq \varnothing$. We first create all possible literals that can be added to the clause given the current body (line $5$). We use \textbf{modes}~\cite{muggleton1994inductive} from inductive logic programming (ILP) to efficiently find the relevant literals here. 

For each literal $l$ in this set, we calculate the true and false groundings for the newly generated clause by adding the literal to the body (line $9$). To perform this calculation, we ground the left hand side of the horn clause (i.e., the query literal) and count the number of groundings of the body corresponding to the query grounding. For instance if the grounding of the query is \emph{advisedBy(John,Mary)} corresponding to \emph{advisedBy(student,prof)}, then we count the number of instances of the body that correspond to {\em John} and {\em Mary}. If the body contains the publications in common, they are counted accordingly. If the body is about courses John took, they are counted correspondingly. This is similar to counting in any relational probabilistic model such as MLNs or BLPs. Following standard SRL models, we assume closed-world. This allows us to deduce the number of false groundings as the difference between the total number of possible groundings and the number of counted (positive) groundings.

We can then calculate the count matrix $C_\ell$ and weights $\bfw$ as described earlier (line $11$--$12$). We score each literal based on the squared error and greedily pick the best scoring literal $\hat{l}$. We repeat this process till the clause reaches its maximum allowed length (set to 4 in our experiments). 

\begin{algorithm}
\begin{algorithmic}[1]
\Function{FitRegression}{$S$, $D$, \pred{y}}
\State where $S$ = $\{\langle y_i, \Delta_i\rangle \}$
\State \pred{body} $\coloneqq \varnothing$; $\bfw \coloneqq [0, 0, 0]^T$ \Comment{Initialize empty clause}
\While {$len(\pred{body}) \le C$}
\State $L$ := \func{PossibleLiterals}(\pred{p}(x), \pred{body}) \Comment{Generate potential literals}
\For {$\ell \in L$} \Comment{Score each literal}
\State \pred{clause} := `\pred{y} $\coloneq$ \pred{body} $\wedge$ $\ell$.'
\For {$x_i \in X$} \Comment{Calculate groundings per example}
\State $t_i, f_i$ = \func{CalculateGroundings}($y_i$, $D$, \pred{clause})
\EndFor
\State $C_l$ := \func{CreateCountMatrix}($\{t_i, f_i\}$)
\State $\bfw(\ell)$ := $\, (C^T C \, + \, \lambda I)^{-1} \, C^T \Delta.$ 
\State $\mathbf{score}(\ell)$ := \func{ScoreFit}($\bfw(\ell), \Delta$)
\EndFor
\State $\hat{\ell} := \arg \min_\ell \mathbf{score}(\ell)$
\State $\bfw := \bfw(\hat{\ell})$
\State \pred{body} := \pred{body} $\wedge$ $\hat{\ell}$
\EndWhile
\State \Return $\bfw$: \pred{y} $\coloneq$ \pred{body}.
\EndFunction
\end{algorithmic}
\caption{Vector-weighted regression clause learning}
\label{algo:fitReg}
\end{algorithm}

To summarize, given a target, the algorithm computes the gradient for all the examples based on the expression $I-P$. Given these gradients, the inner loop searches over the possible clauses such that the MSE is minimized. The resulting vector-weighted clauses are then added to the set of formula and are then used for the subsequent steps of gradient computations. The procedure is repeated until convergence or a preset number of formulae are learned. The search for the most optimal clause can be guided by experts by providing relevant search information as modes~\cite{muggleton1994inductive}. The overall procedure is similar to RDNs and MLNs with two significant differences - the need for multiple weights in the clauses and correspondingly the different optimization function inside the inner loop of the algorithm.

Given that we have outlined the \brlr\, algorithm in detail, we now turn our focus to empirical evaluation of this algorithm.

\section{Experiments and Results}
Our experiments will aim to answer the following questions in order to demonstrate the benefits of \brlr:
\begin{itemize}
    \item[{\bf Q1}] How does functional gradient boosting perform when compared to traditional learning approaches for clauses and weights?
    \item[{\bf Q2}] How does the boosted method perform compared to a significant feature engineered logistic regression approach?
    \item[{\bf Q3}] How does boosting RLR compare to other relational methods?
    \item[{\bf Q4}] How sensitive is the behavior of the proposed approach with respect to the regularization constant, $\lambda$? 
\end{itemize}

\subsection{Methods Considered}
We now compare our $\brlr$ approach to: (1) the $\agglr$ approach, which is standard logistic regression (LR) using the relational information, (2) the $\ilprlr$ approach where rules are learned using a logic learner, followed by weight learning for the formulae, and (3) $\mlnb$, which is a state-of-the-art boosted MLN structure learning method. We evaluate our approach on $1$ synthetic data set and $4$ real world data sets. 
Table~\ref{aggrlr_table} shows the sample aggregate (Relational) features constructed with the highest weights as generated by $\agglr$. 

A natural question to ask is the comparison of our method against the recently successful Boosted Relational Dependency Networks~\cite{natarajan2012gradient} (bRDN) method. We do not consider this comparison for two important reasons - first is that the $\mlnb$ has already been compared against bRDN in the original work and the conclusion was that they were nearly on par in performance in all the domains while bRDN is more efficient due to the use of existentials instead of counts when grounding clauses. Consequently, the second reason is that since our $\agglr$ approach heavily employs counts, we considered the best learning method that employs counts as an aggregator, namely the $\mlnb$ method. Our goal is not to demonstrate that $\agglr$ is more effective than the well-known $\mlnb$ or the bRDN approaches, but to demonstrate that boosting RLR does not sacrifice performance of learners and that RLR can be boosted as effectively as other relational probabilistic models. 

In contrast to our approach, which performs parameter and structure learning {\em simultaneously}, the \ilprlr~baseline performs these steps {\em sequentially}. More specifically, we use PROGOL \cite{Muggleton95,Muggleton97} for structure learning, followed by relational logistic regression for parameter learning. Table \ref{tab: sample rules} shows the number of rules that were learned for each data set by PROGOL. Table \ref{tab: sample rules} also shows some sample rules with the highest coverage scores as generated by PROGOL.

\begin{table}[t]
\centering\setlength{\tabcolsep}{.2\tabcolsep}
\scalebox{0.9}{
\begin{tabular}{|l|c|}
\hline
\multicolumn{1}{|c|}{Domains} & Sample Features \\ \hline
\begin{tabular}[c]{@{}l@{}}$\mathsf{WorkedUnder}$ (IMDB)\\ 5 features constructed\end{tabular} & count\_genres\_acted, count\_movies\_acted \\ \hline
\begin{tabular}[c]{@{}l@{}}$\mathsf{AdvisedBy}$ (UWCS)\\ 8 features constructed\end{tabular} & count\_publications, count\_taughtby \\ \hline
\begin{tabular}[c]{@{}l@{}}$\mathsf{Female}$ (Movie lens)\\ 8 features constructed\end{tabular} & count\_movies, average\_ratings \\ \hline
\begin{tabular}[c]{@{}l@{}}$\mathsf{Cancer}$ (SmCaFr)\\ 3 features constructed\end{tabular} & no\_of\_friends, no\_of\_friends\_smoke \\ \hline
\begin{tabular}[c]{@{}l@{}}$\mathsf{Faculty}$ (WebKB)\\ 4 features constructed\end{tabular} & count\_project, count\_courseta \\ \hline
\end{tabular}
}
\caption{This table shows the number of rules used by AGG-RLR for each data set as well as the features with high weights as picked by LR}
\label{aggrlr_table}
\end{table}

To keep comparisons as fair as possible, we used the following protocol: while employing \mlnb, we set the maximum number of clauses to $3$, the beam-width to $10$ and maximum clause length to $4$. Similar configurations were adopted in our clause-learning setting. Gradient steps for $\mlnb$ and $\brlr$ were picked as per the performance.

\subsection{Data Sets}
\noindent{\bf Smokes-Cancer-Friends}: This is a synthetic data set, where the goal is to predict who has cancer based on the friends network of individuals and their observed smoking habits. The data set has three predicates: \pred{Friends}, \pred{Smokes} and \pred{Cancer}.  We evaluated the method over the \pred{Cancer} predicate using the other predicates with 4-fold cross-validation and $\lambda=10^{3.5}$.\\

\noindent {\bf UW-CSE}: The UW-CSE data set~\cite{richardson2006markov} was created
from the University of Washington’s Computer Science and Engineering
department’s student database and consists of details about professors, students and courses from $5$ different subareas of computer science (AI, programming languages, theory, system and graphics). The data set includes predicates such as \pred{Professor}, \pred{Student}, \pred{Publication}, \pred{AdvisedBy}, \pred{HasPosition},
\pred{TaughtBy} etc., Our task is to learn, using the other predicates, to predict the \pred{AdvisedBy} relation between a student and a professor. There are $4,106,841$ possible \pred{AdvisedBy} relations out of which only $3380$ are true. We employ $5$-fold cross validation where we learn from
four areas and predict on the other area with $\lambda=10^{3.5}$ in our reported results.\\

\noindent{\bf IMDB}: The IMDB data set was first used by Mihalkova and
Mooney~\cite{mihalkova2007bottom} and contains five predicates: \pred{Actor},
\pred{Director},\pred{Movie}, \pred{Genre}, \pred{Gender} and \pred{WorkedUnder}. We predict the \pred{WorkedUnder} relation between an actor and director using the other predicates. Following~\cite{kok2009learning}, we omitted the four equality predicates. We set $\lambda=10^3$ and employed $5$-fold cross-validation using the folds generation strategy suggested by Mihalkova and Mooney in~\cite{mihalkova2007bottom} and averaged the results.\\

\noindent{\bf WebKB}: The WebKB data set was first created by Craven et al.~\cite{craven1998learning} and contains information about department webpages and the links between them. It also contains the categories for each web-page and the words within each page. This data set was converted by Mihalkova and Mooney~\cite{mihalkova2007bottom} to contain only the category of each web-page and links between these pages. They created the following predicates: \pred{Student}, \pred{Faculty}, \pred{CourseTA}, \pred{CourseProf}, \pred{Project} and \pred{SamePerson} from these web-pages. We evaluated the method over the \pred{Faculty} predicate  using the other predicates and we performed $4$-fold cross-validation where each fold corresponds to one university with set $\lambda$ set as $10^2$.\\

\noindent{\bf Movie Lens}: This is the well-known Movielens data set~\cite{HarperKonstan15}
 containing information of about $940$ users, $1682$ movies, the movies rated by each user containing
$79,778$ user-movie pairs, and the actual rating the user has given to a movie. It contains predicates: \pred{Age}, \pred{Genre}, \pred{Occupation}, \pred{Year}, \pred{Ratings} and  \pred{Gender}. In our experiments, we ignored the actual ratings and only considered whether a movie was rated by a user or not. Also, since \pred{Gender} can take only two values, we convert the \pred{Gender}$(person, \, gender)$ predicate to a single argument predicate \pred{FemaleGender}$(person)$. We learned \brlr~models for predicting \pred{FemaleGender} using $5$-fold cross-validation with $\lambda=10^3$.\\

A key property of these relational data sets is the large number of negative examples. This is depicted in Table~\ref{Dataset_table}, which shows the size of various data sets used in our experiments. This is because, in relational settings, a vast majority of relations between objects are not true, and the number of negative examples far outnumbers the number of positive examples. In these data sets, simply measuring accuracy or log-likelihood can be misleading. Hence, we use metrics which are reliable in imbalanced setting like ours.

\begin{table}[]
\centering\setlength{\tabcolsep}{.3\tabcolsep}
\small
\begin{tabular}{c|c|c|c|c}
\hline
\textbf{Data Sets}  & \textbf{Target} & \textbf{Types} & \textbf{Predicates} & \textbf{neg:pos Ratio} \\ \hline
\textbf{Sm-Ca-Fr}  & \pred{Cancer}          & 1              & 3                   & 1.32       \\ 
\textbf{IMDB}      & \pred{WorkedUnder}     & 3              & 6                   & 13.426     \\ 
\textbf{UW-CSE}    & \pred{AdvisedBy}       & 9              & 12                  & 539.629    \\ 
\textbf{WebKB}     & \pred{Faculty}         & 3              & 6                   & 4.159      \\ 
\textbf{Movie Lens} & \pred{FemaleGender}          & 7              & 6             & 2.702      \\ 
\hline
\end{tabular}
\caption{Details of relational domains used in our experiments. These data sets have high ratios of negative to positive examples, which is a key characteristic of relational data sets.}
\label{Dataset_table}
\end{table}

\begin{table*}[t]
\centering
\small
\begin{tabular}{c@{}l}
\hline
\textbf{\small Target (Data Set)} & \textbf{\small Sample rules generated for} \ilprlr~\textbf{using PROGOL} \\ \hline
\begin{tabular}[c]{c} $\mathsf{WorkedUnder}$ (IMDB) \\ $6$ {\em rules generated} \end{tabular} & \begin{tabular}[c]{l@{}l} %
    $\mathsf{WorkedUnder(A, B)} \Leftarrow$ %
    $\mathsf{isa(B, director),\, isa(A, actor),}$%
    $\mathsf{movie(C, A), \, movie(C, B).}$\\ %
    $\mathsf{WorkedUnder(A, B)} \Leftarrow$%
    $\mathsf{genre(B, C), \, gender(A, male).}$ \end{tabular} \\ \hline
    
    
\begin{tabular}[c]{c} $\mathsf{AdvisedBy}$ (UWCS) \\ $16$ {\em rules generated} \end{tabular} & \begin{tabular}[c]{l@{}l} %
    $\mathsf{AdvisedBy(A, B)} \Leftarrow$ $\mathsf{hasPosition(B, E),\, inPhase(A, D),}$ %
    $\mathsf{publication(C, A), \, publication(C, B).}$\\ %
    $\mathsf{AdvisedBy(A, B) \Leftarrow}$ $\mathsf{hasPosition(B, D), \, inPhase(A, E),}$ %
    $\mathsf{publication(F, A).}$ \end{tabular} \\ \hline
    
    
\begin{tabular}[c]{c} $\mathsf{Female}$ (Movie Lens)  \\ $7$ {\em rules generated}  \end{tabular} & \begin{tabular}[c]{l@{}l} %
    $\mathsf{Female(A)} \Leftarrow$ $\mathsf{tmpRatingArg1(B, A), \, tmpRatingArg2(B, C),}$ %
    $\mathsf{genre(C, g4).}$\\ %
    $\mathsf{Female(A)} \Leftarrow$ $\mathsf{age(A, 4), \, occupation(A, o14).}$ \end{tabular} \\ \hline %
    

\begin{tabular}[c]{c} $\mathsf{Cancer}$ (SmCaFr) \\ $3$ {\em rules generated}  \end{tabular} & \begin{tabular}[c]{l@{}l} %
    $\mathsf{Cancer(a) \Leftarrow}$ & $\mathsf{friends(b, a), \, friends(b, c), \, smokes(c).}$\\ %
    $\mathsf{Cancer(a) \Leftarrow}$ & $\mathsf{smokes(a).}$ \end{tabular} \\ \hline %
    

\begin{tabular}[c]{c} $\mathsf{Faculty}$ (WebKB) \\ $6$ {\em rules generated} \end{tabular}  & \begin{tabular}[c]{l@{}l} %
    $\mathsf{Faculty(A) \Leftarrow}$ $\mathsf{courseProf(B, A), \, courseTA(B, C).}$ \\ %
    $\mathsf{Faculty(A) \Leftarrow}$ $\mathsf{courseProf(B, A), \, project(C, A),}$ %
    $\mathsf{samePerson(A, A).}$ \end{tabular} \\ \hline %
\end{tabular}%
\caption{This table shows the number of rules used by \ilprlr~for each data set as well as the rules with the highest $\|P \, - \, N\|$ coverage as returned by PROGOL.}
\label{tab: sample rules}
\end{table*}

\subsection{Results}

\begin{figure}[!t]
    \centering
    \includegraphics[scale =0.18]{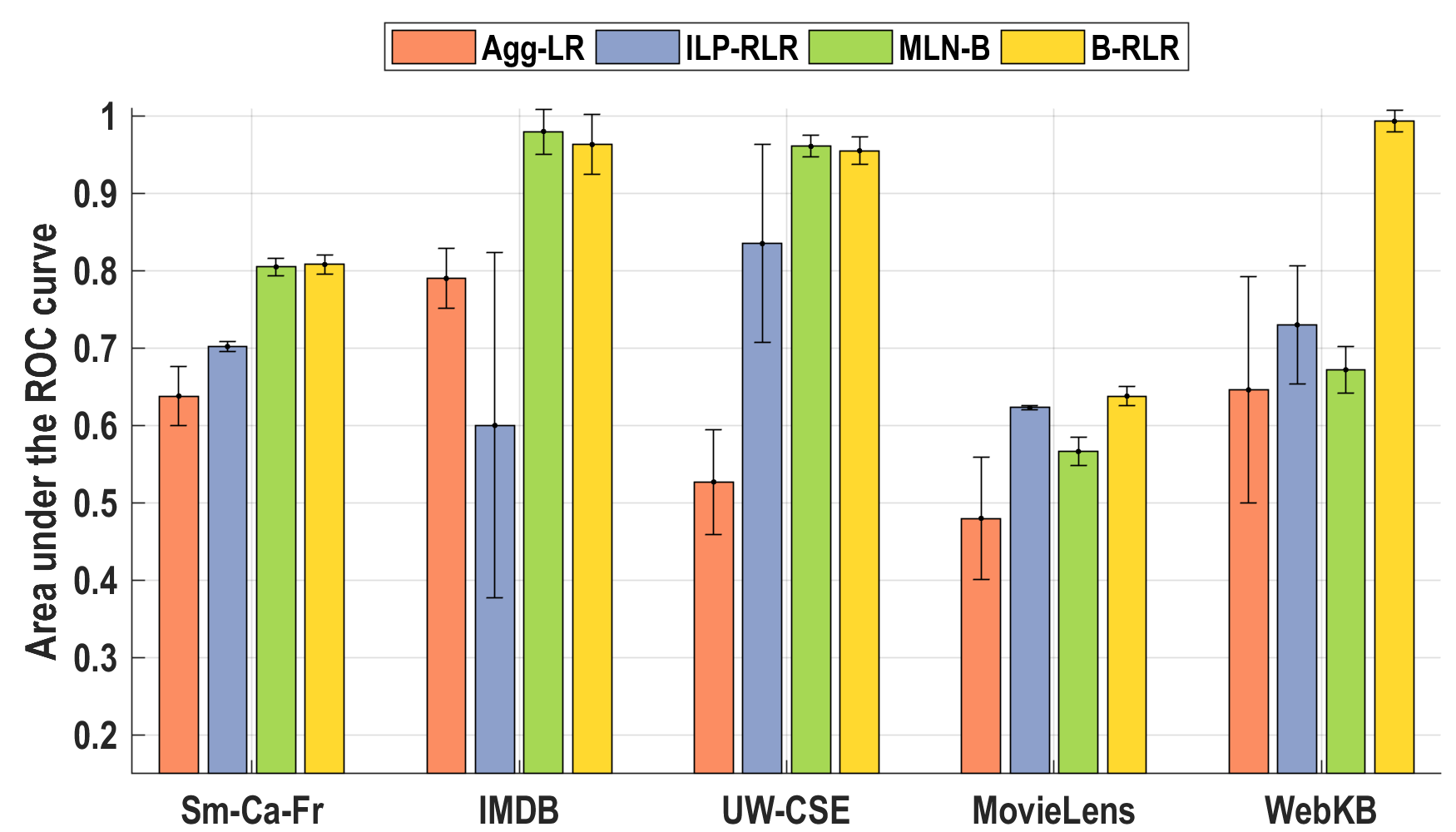}
    \caption{Comparing the area under the ROC curve for the proposed \brlr~approach to (1) standard logistic regression with relational information (\agglr), (2) an approach where rules are learned using a logic learner followed by weight learning (\ilprlr), and (3) state-of-the-art MLN with boosted structure learning (\mlnb).}
    \label{fig:results-aucroc}
\end{figure}

\begin{figure}[!t]
    \centering
    \includegraphics[scale =0.18]{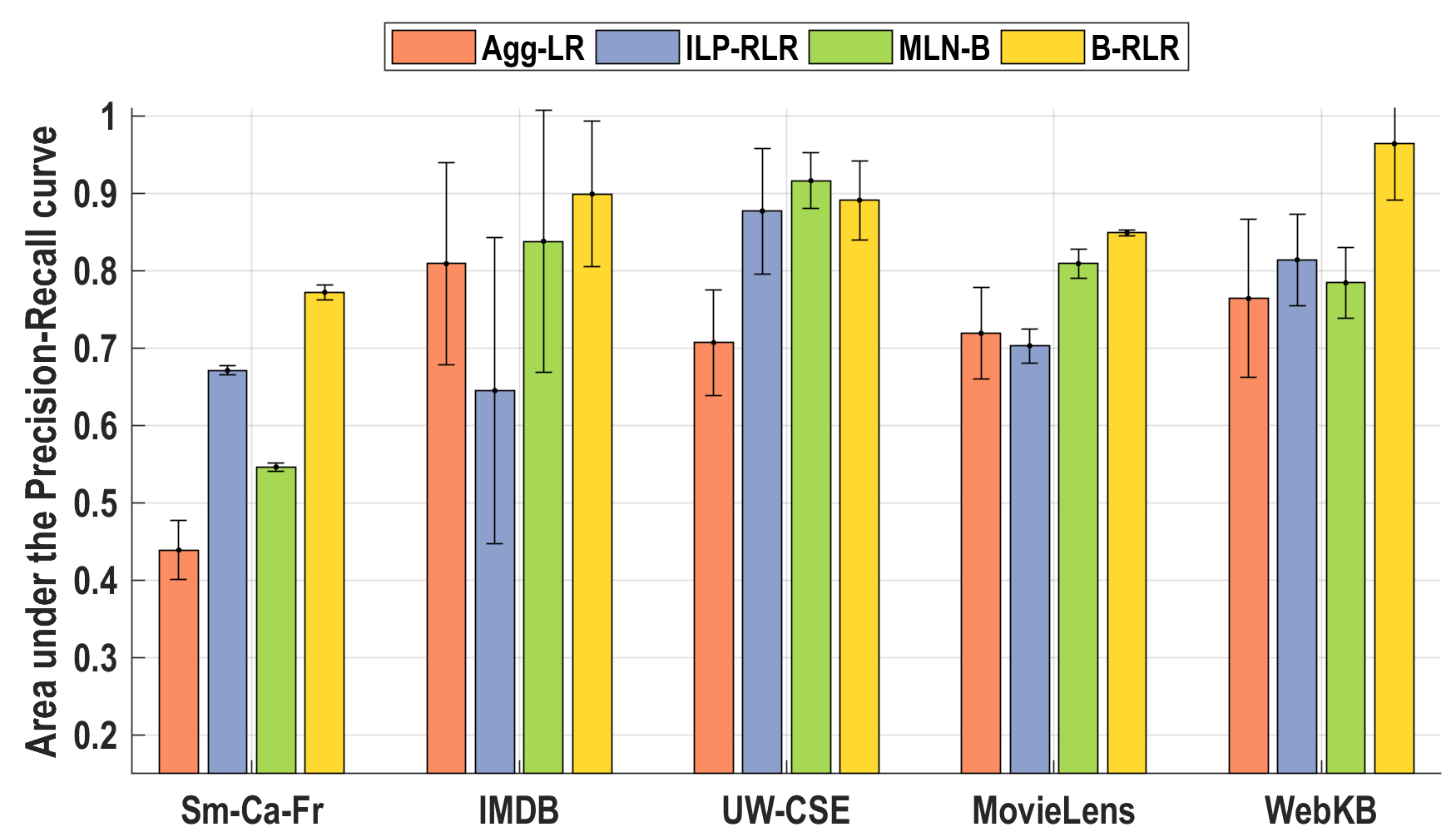}
    \caption{Comparing the area under the Precision-Recall (PR) curve for the proposed \brlr~approach to (1) standard logistic regression with relational information (\agglr), (2) an approach where rules are learned using a logic learner followed by weight learning (\ilprlr), and (3) state-of-the-art MLN with boosted structure learning (\mlnb).}
    \label{fig:results-aucpr}
\end{figure}

We present the results of our experiments in Figures~\ref{fig:results-aucroc} and~\ref{fig:results-aucpr}, which compare the various methods on two metrics: area under the ROC curve (AUC-ROC) and area under the Precision-Recall curve (AUC-PR) respectively. 
From these figures, certain observations can be made clearly. 

First, the proposed \brlr\, method is on par or better than most methods across all data sets. On deeper inspection, it can be seen that the state-of-the-art boosting method for MLNs appears to be more mixed at first glance in ROC-space while \brlr\, is generally better in PR-space. 
In addition, in the WebKB, MovieLens and Smokes-Cancer-Friends domain where we learn about a unary predicate, the performance is significantly better. This yields an interesting insight: RLR models can be \textbf{natural aggregators} over the associated features. As we are in the unary predicate setting (which corresponds to predicting an attribute of an object), the counts of the instances of the body of the clause simply means aggregating over the values of the body. This is typically done in several different ways such as mean, weighted mean  or noisy-or~\cite{Natarajan2008CombiningRules}. We suggest the use of logistic function with counts as an alternative aggregator that seems effective in this domain and we hypothesize its use for many relational tasks where aggregation can yield to natural models. In contrast, MLNs only employ counts as their features, while RLR allows for a more complex aggregation within the sigmoid function that can use count features in its inner loop. Validating this positive aspect of RLR models remains an interesting future research direction. These results help in answering \textbf{Q3} affirmatively: that \brlr\, is on par or significantly better than \mlnb\, in all domains.


\begin{figure}[!t]
    \centering
    \includegraphics[scale =0.18]{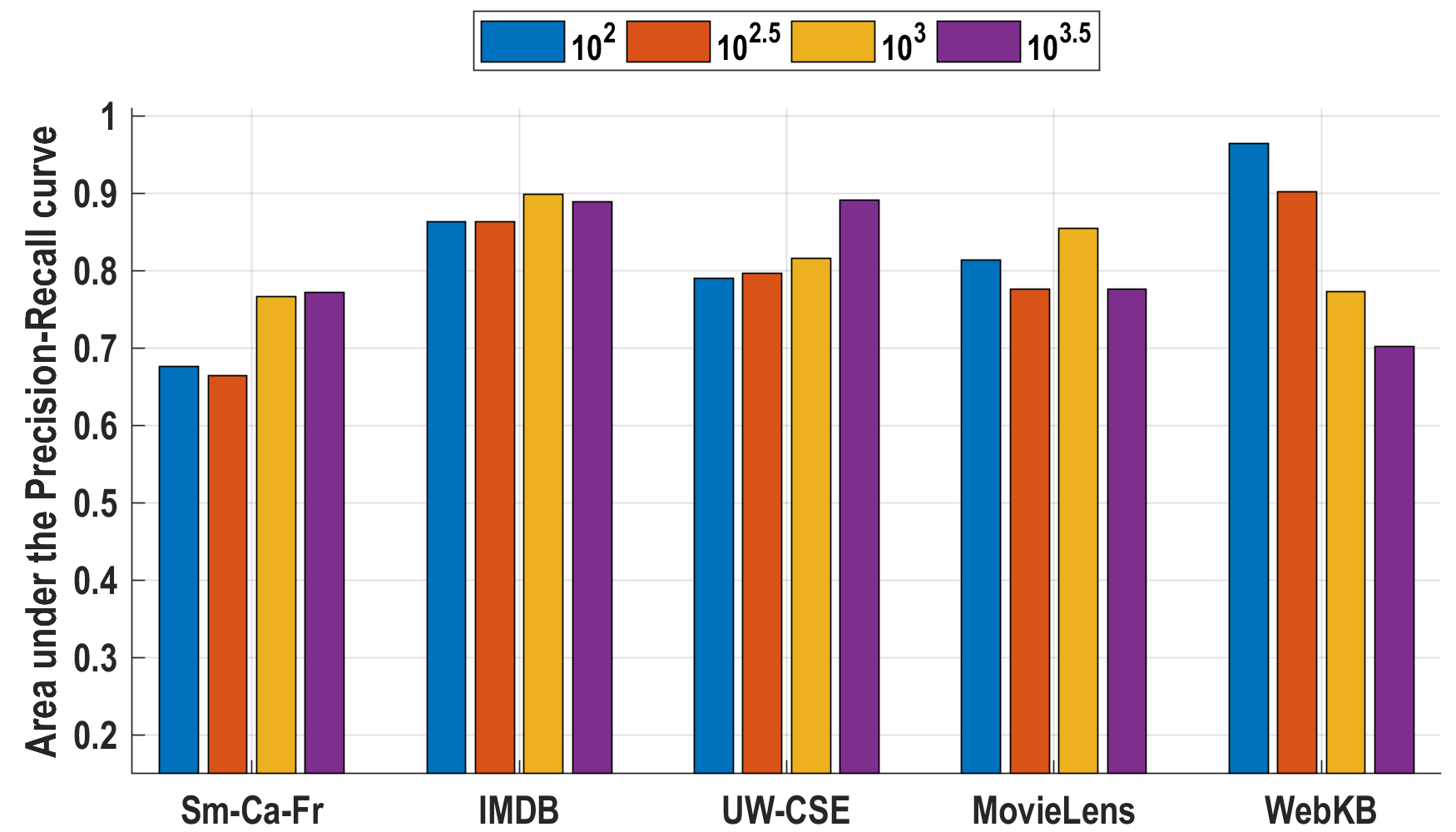}
    \caption{Sensitivity of the proposed \brlr~approach is analyzed by comparing the Area under the Precision-Recall (PR) curve as $\lambda$ changes.}
    \label{fig:results-lamda}
\end{figure}


\begin{table}[t]
\centering\setlength{\tabcolsep}{.3\tabcolsep}
\small
\begin{tabular}{crrrr}
\hline
\multirow{2}{*}{} & \multicolumn{4}{c}{\textbf{Learning Time (seconds)}} \\ \cline{2-5} 
\textbf{Target (Data set)} & \agglr & \ilprlr & \mlnb & \brlr \\ \hline
$\mathsf{WorkedUnder}$ (IMDB) & 67.39 & 158.02 & 6.99 & 10.95 \\
$\mathsf{AdvisedBy}$ (UWCS) & 110.58 & 65.058 & 18.40 & 24.71 \\ 
$\mathsf{Female}$ (Movie Lens) & 91.27 & 78.57 & 6.66 & 16.51 \\
$\mathsf{Cancer}$ (SmCaFr) & 35.18 & 3.26 & 120.73 & 90.47 \\ 
$\mathsf{Faculty}$ (WebKB) & 51.73 & 2.18 & 5.68 & 5.86 \\\hline
\end{tabular}
\caption{Comparing learning time (in seconds)  for  the  proposed \brlr~approach to (1) standard logistic regression with relational information (\agglr), (2) an approach where rules are learned using a logic learner followed by weight learning (\ilprlr), and (3) state-of-the-art MLN with boosted structure learning (\mlnb). Learning time includes time to learn the structure, counting the satisfied (or unsatisfied) groundings and weight learning.}
\label{tab: running times}
\hspace{-0.25in}
\end{table}

Next, from the figures, it can be observed that \brlr\, significantly outperforms \agglr\, in several domains. At the outset, this may not be surprising since relational models have been shown to outperform non-relational models. However, the features that are created for the \agglr\, model are the count features of the type defined in the original RLR work and are more expressive than the standard features of propositional models. This result is particularly insightful as the \brlr\, model that uses count features, predicates and their combinations themselves in a formula is far more expressive than simple aggregate features. This allows us to answer {\bf Q2} strongly and affirmatively, in that the proposed approach is significantly better than an engineered (relational) logistic regression approach.

Finally, comparing the proposed approach to a two-step approach of learning clauses followed by the corresponding weights; \brlr\, appears to be significantly better in both PR-space as well as ROC-space than \ilprlr\,. Furthermore, \brlr\, also has the distinct advantage of simultaneous parameter and structure learning, thus avoiding a costly structure search compared to the ILP-based approach. Hence, {\bf Q1} can be answered: that \brlr\, model outperforms ILP-based two-stage learning in all the regions. We see from Figure~\ref{fig:results-lamda} that {\bf Q4} is answered affirmatively: As long as $\lambda$ is within the set $\{10^2, 10^{2.5}, 10^3, 10^{3.5}\}$, our algorithm is not sensitive in most domains. In addition, our parameter $\lambda$ has a nice intuitive interpretation: they reflect and incorporate the high class imbalance that exists for real-world domains where this is an important practical consideration. 

We used paired t-test with p-values=0.05 for determining the statistical significance. From the Figures~\ref{fig:results-aucroc} \&~\ref{fig:results-aucpr}, across most domains, we observe that, \brlr\, has tighter error bounds compared to the baselines in majority of domains indicating lower variance and subsequently higher generalization performance. 

Table~\ref{tab: running times} reports the training time taken in seconds by each method averaged over all the folds in every domain. Timings reported For \agglr\, include time taken for propositional feature construction and weight learning using WEKA tool. For \ilprlr\, it includes the total time taken to learn rules, count satisfied instances, and learn weights for the rules accordingly. The two boosted approaches timings are reported from the full runs. The results show that the methods are comparable across all the domains - in the domains where boosted methods are faster than the other baselines, grounding of the entire data set caused the increased time for the baselines. Conversely, in the other domains, repeated counting of boosting increased the time in two of the five domains. The results indicate that the proposed \brlr approach does not sacrifice efficiency (time) for effectiveness (performance). 

In summary, our proposed boosted approach appears to be promising across all a diversity of relational domains, with the potential for scaling up relational logistic regression models to large data sets. 

\section{Conclusions}

We considered the problem of learning relational logistic regression (RLR) models using the machinery of functional-gradient boosting. To this end, we introduce an alternative interpretation of RLR models that allows us to consider both the true and false groundings of a formula within a single equation. This allowed us to learn vector-weighted clauses that are more compact and expressive compared to standard boosted SRL models. We derived gradients for the different weights, and outlined a learning algorithm that learned first-order features as clauses and the corresponding weights simultaneously. We evaluated the algorithm on standard data sets, and demonstrated the efficacy of the learning algorithm.

There are several possible extensions for future work -- currently, our method learns a model for a single target predicate deterministically. As mentioned earlier, it is possible to learn a joint model across multiple predicates in a manner akin to learning a relational dependency network (RDN). This can yield a new interpretation for RDNs based on combining rules. Second, learning from truly hybrid data remains a challenge for SRL models in general, and RFGB in particular. Finally, given the recent surge of sequential decision-making research, RLR can be seen as an effective function approximator for relational Markov decision processes (MDPs); employing this novel \brlr\, model in the context of relational reinforcement learning can be an exciting and interesting future research direction.

\bibliographystyle{aaai}
\bibliography{rlrbib}

\begin{thebibliography}{}

\bibitem[\protect\citeauthoryear{Blockeel}{1999}]{tilde}
Blockeel, H.
\newblock 1999.
\newblock Top-down induction of first order logical decision trees.
\newblock {\em AIC} 12(1-2).

\bibitem[\protect\citeauthoryear{Craven \bgroup et al\mbox.\egroup
  }{1998}]{craven1998learning}
Craven, M.; DiPasquo, D.; Freitag, D.; McCallum, A.; Mitchell, T.; Nigam, K.;
  and Slattery, S.
\newblock 1998.
\newblock Learning to extract symbolic knowledge from the world wide web.
\newblock In {\em AAAI},  509--516.

\bibitem[\protect\citeauthoryear{Dietterich, Ashenfelter, and
  Bulatov}{2004}]{dietterich04}
Dietterich, T.; Ashenfelter, A.; and Bulatov, Y.
\newblock 2004.
\newblock Training {CRFs} via gradient tree boosting.
\newblock In {\em ICML}.

\bibitem[\protect\citeauthoryear{Fatemi, Kazemi, and
  Poole}{2016}]{FatemiKP16-LearningRLR}
Fatemi, B.; Kazemi, S.~M.; and Poole, D.
\newblock 2016.
\newblock A learning algorithm for relational logistic regression: Preliminary
  results.
\newblock {\em arXiv preprint arXiv:1606.08531}.

\bibitem[\protect\citeauthoryear{Friedman}{2001}]{friedman01}
Friedman, J.
\newblock 2001.
\newblock Greedy function approximation: A gradient boosting machine.
\newblock {\em Annals of Statistics} 29.

\bibitem[\protect\citeauthoryear{Getoor and Taskar}{2007}]{GetoorTaskar07}
Getoor, L., and Taskar, B.
\newblock 2007.
\newblock {\em Introduction to Statistical Relational Learning}.
\newblock MIT Press.

\bibitem[\protect\citeauthoryear{Harper and Konstan}{2015}]{HarperKonstan15}
Harper, F., and Konstan, J.
\newblock 2015.
\newblock The {MovieLens} datasets: History and context.
\newblock {\em ACM Trans. Interact. Intell. Syst.} 5(4):19:1--19:19.

\bibitem[\protect\citeauthoryear{Heckerman, Meek, and
  Koller}{2007}]{heckerman2007probabilistic}
Heckerman, D.; Meek, C.; and Koller, D.
\newblock 2007.
\newblock Probabilistic entity-relationship models, {PRMs}, and plate models.
\newblock {\em Introduction to statistical relational learning}  201--238.

\bibitem[\protect\citeauthoryear{Huynh and
  Mooney}{2008}]{huynh2008discriminative}
Huynh, T., and Mooney, R.
\newblock 2008.
\newblock Discriminative structure and parameter learning for {Markov} logic
  networks.
\newblock In {\em ICML},  416--423.
\newblock ACM.

\bibitem[\protect\citeauthoryear{Kazemi and Poole}{2018}]{relnn}
Kazemi, S.~M., and Poole, D.
\newblock 2018.
\newblock {RelNN}: A deep neural model for relational learning.
\newblock In {\em AAAI}.

\bibitem[\protect\citeauthoryear{Kazemi \bgroup et al\mbox.\egroup
  }{2014a}]{rlr}
Kazemi, S.; Buchman, D.; Kersting, K.; Natarajan, S.; and Poole, D.
\newblock 2014a.
\newblock Relational logistic regression.
\newblock In {\em KR}.

\bibitem[\protect\citeauthoryear{Kazemi \bgroup et al\mbox.\egroup
  }{2014b}]{KazemiEtAl14-RLR}
Kazemi, S.; Buchman, D.; Kersting, K.; Natarajan, S.; and Poole, D.
\newblock 2014b.
\newblock Relational logistic regression: The directed analog of markov logic
  networks.
\newblock In {\em AAAI Workshop}.

\bibitem[\protect\citeauthoryear{Kersting and
  De~Raedt}{2007}]{KerstingDeRaedt07-BLP}
Kersting, K., and De~Raedt, L.
\newblock 2007.
\newblock Bayesian logic programming: {T}heory and tool.
\newblock In Getoor, L., and Taskar, B., eds., {\em An Introduction to
  Statistical Relational Learning}.

\bibitem[\protect\citeauthoryear{Khot \bgroup et al\mbox.\egroup
  }{2011}]{khot2011learning}
Khot, T.; Natarajan, S.; Kersting, K.; and Shavlik, J.
\newblock 2011.
\newblock Learning {Markov} logic networks via functional gradient boosting.
\newblock In {\em ICDM},  320--329.
\newblock IEEE.

\bibitem[\protect\citeauthoryear{Kimmig \bgroup et al\mbox.\egroup
  }{2012}]{kimmig2012short}
Kimmig, A.; Bach, S.; Broecheler, M.; Huang, B.; and Getoor, L.
\newblock 2012.
\newblock A short introduction to probabilistic soft logic.
\newblock In {\em NIPS Workshop},  1--4.

\bibitem[\protect\citeauthoryear{Kok and Domingos}{2009}]{kok2009learning}
Kok, S., and Domingos, P.
\newblock 2009.
\newblock Learning {Markov} logic network structure via hypergraph lifting.
\newblock In {\em ICML},  505--512.
\newblock ACM.

\bibitem[\protect\citeauthoryear{Koller}{1999}]{koller1999probabilistic}
Koller, D.
\newblock 1999.
\newblock Probabilistic relational models.
\newblock In {\em ILP},  3--13.
\newblock Springer.

\bibitem[\protect\citeauthoryear{Malec \bgroup et al\mbox.\egroup
  }{2016}]{malec2016inductive}
Malec, M.; Khot, T.; Nagy, J.; Blasch, E.; and Natarajan, S.
\newblock 2016.
\newblock Inductive logic programming meets relational databases: An
  application to statistical relational learning.
\newblock In {\em ILP}.

\bibitem[\protect\citeauthoryear{McCullagh}{1984}]{mccullagh1984generalized}
McCullagh, P.
\newblock 1984.
\newblock Generalized linear models.
\newblock {\em EJOR} 16(3):285--292.

\bibitem[\protect\citeauthoryear{Mihalkova and
  Mooney}{2007}]{mihalkova2007bottom}
Mihalkova, L., and Mooney, R.
\newblock 2007.
\newblock Bottom-up learning of {Markov} logic network structure.
\newblock In {\em ICML},  625--632.
\newblock ACM.

\bibitem[\protect\citeauthoryear{Muggleton and
  De~Raedt}{1994}]{muggleton1994inductive}
Muggleton, S., and De~Raedt, L.
\newblock 1994.
\newblock Inductive logic programming: Theory and methods.
\newblock {\em Journal of Logic Programming} 19:629--679.

\bibitem[\protect\citeauthoryear{Muggleton}{1995}]{Muggleton95}
Muggleton, S.
\newblock 1995.
\newblock Inverse entailment and progol.
\newblock {\em New generation computing} 13(3-4):245--286.

\bibitem[\protect\citeauthoryear{Muggleton}{1997}]{Muggleton97}
Muggleton, S.
\newblock 1997.
\newblock Learning from positive data.
\newblock In {\em ILP},  358--376.
\newblock Berlin, Heidelberg: Springer Berlin Heidelberg.

\bibitem[\protect\citeauthoryear{Natarajan \bgroup et al\mbox.\egroup
  }{2008}]{Natarajan2008CombiningRules}
Natarajan, S.; Tadepalli, P.; Dietterich, T.; and Fern, A.
\newblock 2008.
\newblock Learning first-order probabilistic models with combining rules.
\newblock {\em Annals of Mathematics and Artificial Intelligence}
  54(1-3):223--256.

\bibitem[\protect\citeauthoryear{Natarajan \bgroup et al\mbox.\egroup
  }{2012}]{natarajan2012gradient}
Natarajan, S.; Khot, T.; Kersting, K.; Gutmann, B.; and Shavlik, J.
\newblock 2012.
\newblock Gradient-based boosting for statistical relational learning: The
  relational dependency network case.
\newblock {\em Machine Learning} 86(1):25--56.

\bibitem[\protect\citeauthoryear{Neville and
  Jensen}{2007}]{NevilleJensen07-RDN}
Neville, J., and Jensen, D.
\newblock 2007.
\newblock Relational dependency networks.
\newblock In Getoor, L., and Taskar, B., eds., {\em Introduction to Statistical
  Relational Learning}. MIT Press.
\newblock  653--692.

\bibitem[\protect\citeauthoryear{Poole \bgroup et al\mbox.\egroup
  }{2014}]{Poole2014}
Poole, D.; Buchman, D.; Kazemi, S.; Kersting, K.; and Natarajan, S.
\newblock 2014.
\newblock Population size extrapolation in relational probabilistic modelling.
\newblock In {\em SUM},  292--305.
\newblock Springer.

\bibitem[\protect\citeauthoryear{Raedt \bgroup et al\mbox.\egroup
  }{2016}]{DeRaedtEtAl16}
Raedt, L.; Kersting, K.; Natarajan, S.; and Poole, D.
\newblock 2016.
\newblock Statistical relational artificial intelligence: Logic, probability,
  and computation.
\newblock {\em Synthesis Lectures on AI and ML} 10(2):1--189.

\bibitem[\protect\citeauthoryear{Richardson and
  Domingos}{2006}]{richardson2006markov}
Richardson, M., and Domingos, P.
\newblock 2006.
\newblock {Markov} logic networks.
\newblock {\em ML} 62(1):107--136.

\bibitem[\protect\citeauthoryear{Taskar \bgroup et al\mbox.\egroup
  }{2007}]{taskar2007relational}
Taskar, B.; Abbeel, P.; Wong, M.; and Koller, D.
\newblock 2007.
\newblock Relational {Markov} networks.
\newblock {\em Introduction to statistical relational learning}  175--200.

\bibitem[\protect\citeauthoryear{Yang \bgroup et al\mbox.\egroup
  }{2017}]{yang2017combining}
Yang, S.; Korayem, M.; AlJadda, K.; Grainger, T.; and Natarajan, S.
\newblock 2017.
\newblock Combining content-based and collaborative filtering for job
  recommendation system: A cost-sensitive statistical relational learning
  approach.
\newblock {\em KBS} 136:37--45.

\end{thebibliography}

\end{document}